\documentclass[10pt,twocolumn,letterpaper]{article}

\usepackage{iccv}
\usepackage{times}
\usepackage{epsfig}
\usepackage{graphicx}
\usepackage{amsmath}
\usepackage{amssymb}
\usepackage[bottom]{footmisc}

\usepackage{booktabs}
\usepackage[table]{xcolor}

\definecolor{yelloworange}{RGB}{255, 153, 0}
\definecolor{ultramarineblue}{RGB}{65, 102, 245}
\usepackage{comment}
\usepackage[pagebackref=true,breaklinks=true,letterpaper=true,colorlinks,bookmarks=false]{hyperref}

\iccvfinalcopy 

\def\iccvPaperID{49} 

\ificcvfinal\fi
\begin{document}

\title{Recognizing Part Attributes with Insufficient Data}

\author{Xiangyun Zhao\\
{Northwestern University}
\and
Yi Yang\\
{Baidu Research}
\and 
Feng Zhou \\
{Baidu Research}
\and 
Xiao Tan \\
{Baidu Inc.}
\and 
Yuchen Yuan \\
{Baidu Inc.}
\and 
Yingze Bao \\
{Baidu Research}
\and
Ying Wu \\
{Northwestern University}
}

\maketitle

\begin{abstract}

Recognizing attributes of objects and their parts is important to many computer vision applications. 
Although great progress has been made to apply object-level recognition, recognizing the attributes of parts remains less applicable since the training data for part attributes recognition is usually scarce especially for internet-scale applications. 
Furthermore, most existing part attribute recognition methods rely on the part annotation which is more expensive to obtain. 
To solve the data insufficiency problem and get rid of dependence on the part annotation, we introduce a novel Concept Sharing Network (CSN) for part attribute recognition. 
A great advantage of CSN is its capability of recognizing the part attribute (a combination of part location and appearance pattern) that has insufficient or zero training data, by learning the part location and appearance pattern respectively from the training data that usually mix them in a single label. 
Extensive experiments on CUB-200-2011~\cite{wah2011caltech}, CelebA~\cite{liu2015deep} and a newly proposed human attribute dataset demonstrate the effectiveness of CSN and its advantages over other methods, especially for the attributes with few training samples. 
Further experiments show that CSN can also perform zero-shot part attribute recognition. The code will be made available at https://github.com/Zhaoxiangyun/Concept-Sharing-Network.



\end{abstract}

\section{Introduction}

The computer vision community has seen tremendous progress in recognizing global features of objects, such as performing category detection~\cite{ren2015faster,girshick2015fast,redmon2016you,zhao2018pseudo} and classification~\cite{krizhevsky2012imagenet} (e.g. detect the bounding box and classify the category of a bird from an image).
Meanwhile, recognizing attributes of object parts (e.g. localize the wing of a bird and classify its biologic feature) is still a very challenging problem due to multiple issues. First, attributes (e.g. the color of the wing of a bird ) normally attach to a very limited area of an object, which are usually more difficult to be accurately localized from an image compared to the overall object. Most existing part attribute recognition methods~\cite{zhang2014panda, li2016human, zhang2014part} train a part detector with large extra annotations to detect the relevant part. However, such part annotations are very expensive to obtain. Therefore, these methods generally fail when the part annotation is not available. How to recognize the part attribute with only image-level annotation is still under-explored. 

\begin{figure}[t]
	\centering
	\includegraphics[width=0.9\linewidth]{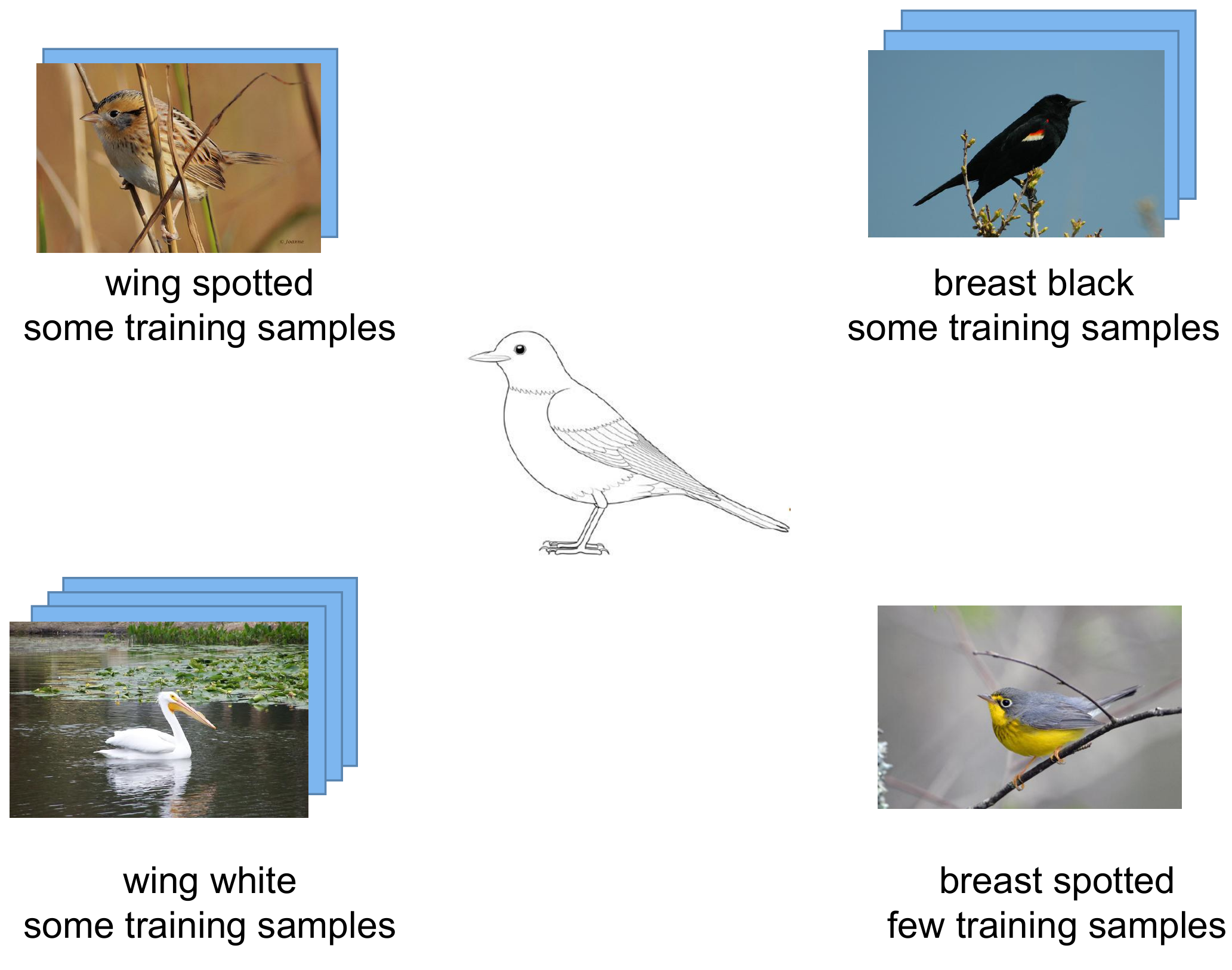}
	\caption{In many datasets and real applications, the labeling of part attributes is usually very limited.
	For example, as this figure shows, in CUB-200-2011~\cite{wah2011caltech} dataset the labels of \textit{breast spotted} is very few, whereas the number of the labels of \textit{wing spotted}, \textit{wing white}, and \textit{breast black} are more but still limited. We propose to identify the relationships between different labels based on their locations and patterns, so as to re-use the labels of other attributes to facilitate the recognition of the attribute that lacks labels (e.g. \textit{breast spotted} in this figure). Further, we find that the recognition of all these attributes can be jointly improved if individual concepts of different attributes can be shared.}
	\label{fig.motivation}
	\vspace{-3mm}
\end{figure}

Another important problem is that the training data is expensive to obtain and usually insufficient in the existing dataset.
For example, in a commonly used bird parts attribute recognition dataset CUB-200-2011~\cite{wah2011caltech}, the number of training images for most attributes varies from merely a few dozens to at most a few hundred.
Most existing attribute recognition methods (if not all) process each part attribute independently, and ignore the spatial correlation of different part attributes. 
As a result, their performance is simply limited by the volume of training data of each isolated attribute. How to solve the data insufficiency problem is rarely discussed.

\begin{figure}
	\centering
	\includegraphics[width=\linewidth]{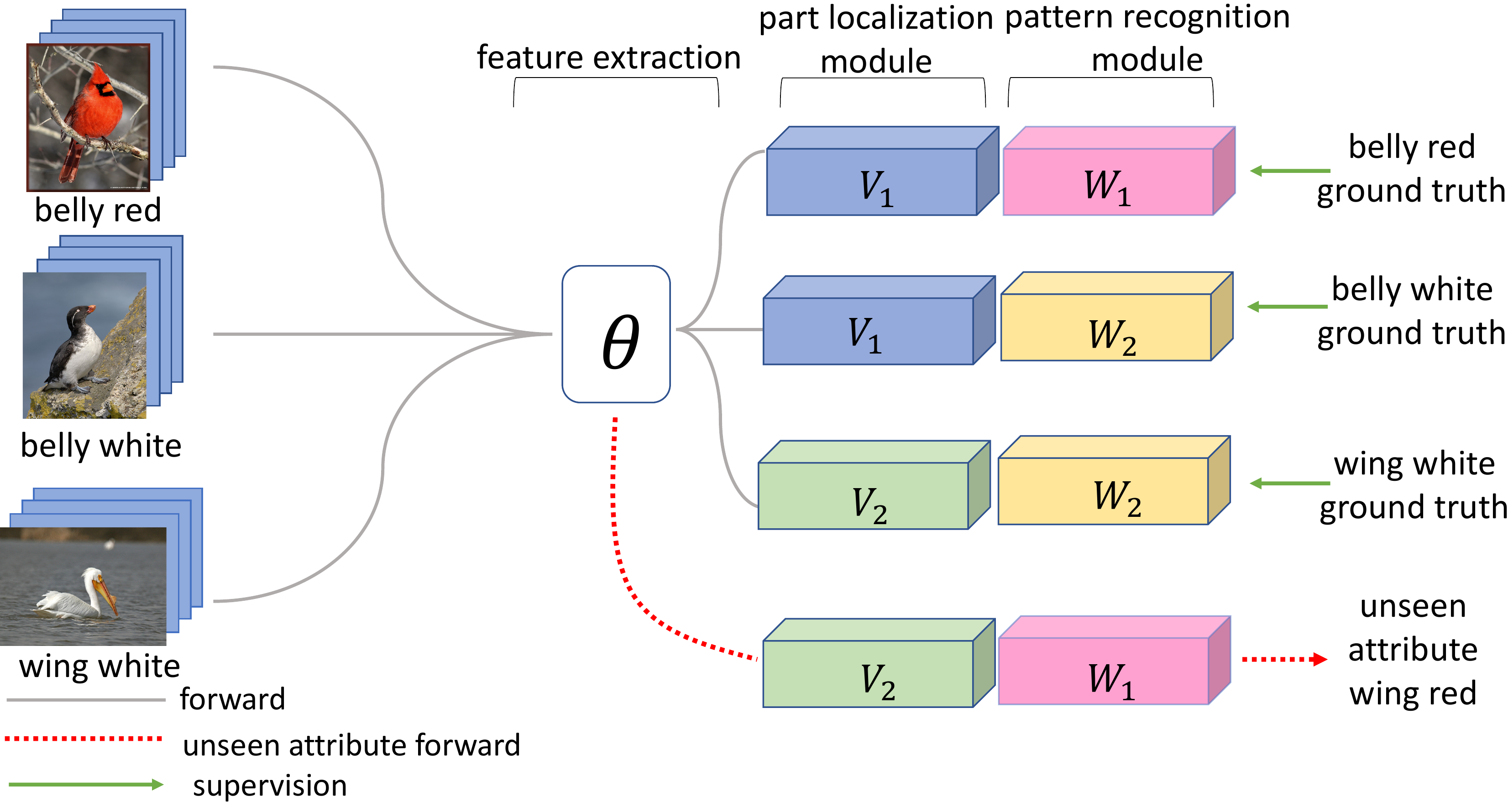}
	\caption{\textbf{Overview of the training.} Training images of different attributes are forwarded through the CNN to obtain the image representation, then attribute samples with the same part are forwarded through the same localization module and attribute samples with the same appearance pattern are forwarded through the same pattern recognition module. New attribute with no training data could be recognized as the combination of the learned modules.}
	\label{fig.framework1}
	\vspace{-3mm}
\end{figure}

To address these challenges, we propose a novel Concept Sharing Network (CSN) for part attribute recognition.
In CSN, the part attribute is defined as the combination of two concepts: part location and appearance pattern, as illustrated in Fig.~\ref{fig.motivation}. 
Our neural network models the two concepts as two modules, in contrast to individually modeling each attribute in different branches.
Since the two modules in CSN could be shared among different attributes, the labeling of attributes (e.g. color and shape) belong to certain parts (e.g. wing) of an object can be used to facilitate the training of another attribute of the same parts, and vice versa.
In such a manner, we maximize the usage of the precious training data to boost the attribute recognition performance independently and aggregately. Note that CSN only needs image-level attributes labels to train, so it would be more general than existing part attribute recognition methods~\cite{zhang2014panda, li2016human, zhang2014part} which depend on the part location annotation. 

Furthermore, CSN can be used to discover new attributes, \ie zero-shot part attribute recognition.
Given a training set with certain attributes, the training result of part localization and pattern recognition in CSN could be used to recognize a new attribute that does not belong to the training set.
As illustrated in Fig.~\ref{fig.framework1}, after the \textit{wing} location and color (\textit{red}) pattern are learned, a new attribute \textit{wing red} could be recognized even though the training data of such attribute does not exist. 

In this work, we also contribute a large-scale human attribute dataset, named as SurveilA, which contains 75,000 images with 10 carefully annotated attributes focusing on the fine-grained human activities for video surveillance.
The human images are collected in the wild under different scenes, scales, poses and viewpoint variations.
The dataset is challenging as shown in the experiments that simply fine-tuning standard networks cannot provide accurate enough estimation, and recognition would require a model to focus on local discriminative parts.

Overall, our work has the following contributions: 
1) We aim at addressing the data insufficiency problem in part attribute recognition, which is rarely discussed in previous work.
2) We present an effective part attribute recognition framework which does not depend on the part annotation.
3) Our network is proven to be effective in zero-shot part attribute recognition.
4) We will release a new dataset for part attribute recognition, which consists of 75000 images of human in a real-world scenario with 10 attributes annotation.

\section{Related Work}

\textbf{Attribute Recognition}
Attribute Recognition was first introduced as a computer vision problem in~\cite{ferrari2008learning}. 
From then, attribute recognition have been studied extensively with numerous datasets and methods~\cite{farhadi2009describing,farhadi2010attribute,kumar2009attribute,kumar2011describable,lampert2009learning,lampert2014attribute,wang2017attribute, zhao2018modulation}.
Part Attribute recognition is a harder problem because it is only attached to a very limited area of an object. 
State of the art methods~\cite{bourdev2011describing, zhang2014panda,li2016human} usually rely on the part location annotations to train part detectors such as Poselets~\cite{bourdev2011describing}, Deformable Part Models~\cite{zhang2013deformable} or R-CNN~\cite{girshick2016region} to first localize parts then extract visual features to recognize attributes~\cite{joo2013human}. But the part annotation is very expensive to obtain. Although, recently some methods~\cite{Woo_2018_ECCV,jaderberg2015spatial, zhou2016learning} are proposed to localize the important regions for recognition, they are not carefully designed for part attribute recognition and do not attempt to address the data insufficiency problem. ~\cite{gebru2017fine,liu2017localizing} used attribute recognition results to facilitate the fine-grained recognition, but both of them will fail when training data is insufficient.

\textbf{Few-shot / Zero-shot Learning}
Besides collecting more data, few-shot learning~\cite{thrun1996learning} and zero-shot learning~\cite{palatucci2009zero} attempt to directly address the data insufficiency problem - predicting novel concepts that were either very few or completely unseen from the training set.
These problems are classical because almost all in-the-wild data follow a heavy-tail distribution~\cite{hill1975simple} with new classes appearing frequently after the training and no finite set of samples can cover the diversity of the real world.
Recently, few-shot learning is modeled as a meta learning problem~\cite{ravi2016optimization,finn2017model} through explicitly building training loss to enforce adaptation to new categories with few examples.
On the other hand, due to the complete lack of training data, zero-shot models attempt to learn to
transfer knowledge from other external sources~\cite{al2015transfer,romera2015embarrassingly,chao2016empirical,wang2016relational,zhang2016zero,li2017zero,xian2017zero}.
In contrast to these works, we make use of the visual attention mechanism to disentangle part location with appearance features and share the disentangled representations between attributes, which enable us to conduct zero-shot or few-shot generalization on novel attributes.

\textbf{Visual Attention and Visual Question Answering}
Visual attention models~\cite{mnih2014recurrent,ba2014multiple} have been widely used in object recognition~\cite{zhou2016learning,wang2017residual}, fine-grained recognition~\cite{sermanet2014attention,liu2016fully}, image captioning~\cite{xu2015show} and visual question answering~\cite{chen2015abc}.
These models also decompose the location and appearance during representation, but do not focus on addressing the data insufficiency problem.
CSN improves the visual attention models on data efficiency by sharing the attention mechanism across multiple attribute labels.

The attention region and feature sharing depend on attribute labels, which is similar to visual question answering (VQA)~\cite{antol2015vqa}.
In VQA, existing methods~\cite{andreas2016neural,mascharka2018transparency} also try to localize the relevant region according to the given question. 
But in VQA, all Q/A pairs use the same classifier (i.e. sharing image and language feature extraction, sharing answer classifier). 
In contrast, recognition of multiple attributes are usually considered as a multi-task problem with different classifiers to be trained.
It is very expensive to collect sufficient data so as to well train each classifier, especially in large-scale recognition.
This poses a special challenge of data insufficiency, which is the focus of our method. 

\textbf{Attribute Dataset}
There are some attribute recognition datasets existing, from general objects and scenes~\cite{farhadi2009describing,lampert2009learning,farhadi2010attribute,russakovsky2010attribute,patterson2012sun,lampert2014attribute,patterson2016coco,zhao2018large} to specific fine-grained classes such as faces~\cite{huang2008labeled,kumar2009attribute,kumar2011describable,liu2015deep}, birds~\cite{wah2011caltech}, cars~\cite{yang2015large}, clothes~\cite{liu2016deepfashion} or even butterflies~\cite{wang2009learning}. Due to the importance of human attribute recognition~\cite{gray2007evaluating,bourdev2011describing,sharma2011learning,zhu2013pedestrian,deng2014pedestrian,zhang2014panda,sudowe2015person,li2016richly,li2016human} in real world applications and challenges, we propose a large scale challenging dataset for human part attribute recognition.


\section{Concept Sharing Network}
\begin{figure*}[t]
	\centering
	\includegraphics[width=0.86\linewidth]{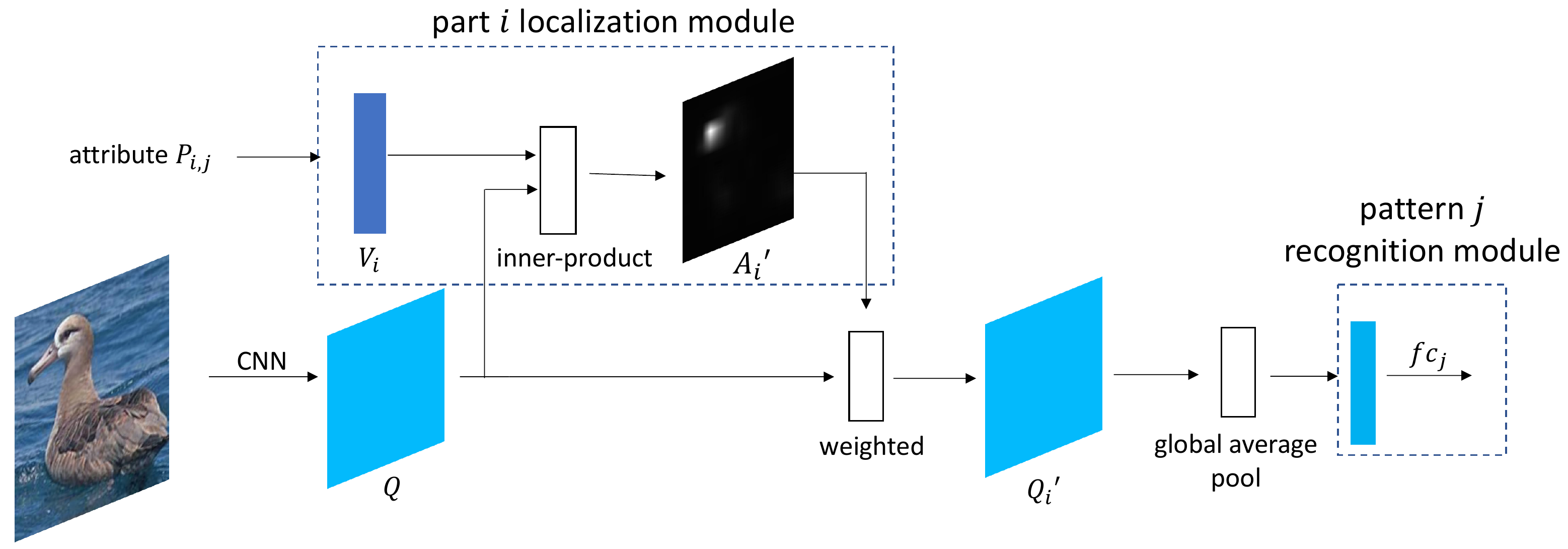}
	\caption{\textbf{Overview of the inference.} In the inference, given an image and recognizing attribute $P_{i,j}$, the image is forwarded through the corresponding location module $i$ and pattern module $j$ to obtain the final prediction. }
	\label{fig.framework2}
	\vspace{-3mm}
\end{figure*}
Part attribute recognition aims to predict whether or not the statement of the attribute of a part is true, e.g. whether or not 'the wing of the bird is black' or 'the bill of the bird is red'.
In this work, we introduce a novel Concept Sharing Network(CSN) for part attribute recognition, as illustrated in Fig. \ref{fig.framework2}.
In CSN, each attribute is defined as a combination of two concepts: part localization and pattern recognition.
\subsection{Network}
In CSN, attributes are recognized based on two modules: the part localization module and the appearance pattern recognition module.
In our training process, as illustrated in Fig.~\ref{fig.framework1}, training images of different attributes are forwarded through the CNN to obtain the image representation, then attribute samples with the same part are forwarded through the same localization module and attribute samples with the same appearance pattern are forwarded through the same pattern recognition module.
The modules are learned by the training samples which are forwarded through them.
In the inference process, as illustrated in Fig.~\ref{fig.framework2}, given an image and recognizing attribute $P_{i,j}$ corresponding to part $i$ and appearance pattern $j$, the image is forwarded through the corresponding location module and pattern module to obtain the final prediction.
\subsubsection{Part Localization module}
One novelty of our work is the employment of attention mechanism within CSN neural network to localize the part for attributes.
We propose an attention based method inspired by CAM~\cite{zhou2016learning}.
Note that many other alternatives~\cite{wang2017residual,Wang_nonlocalCVPR2018} can also be incorporated with our framework.
Given an image $x$, we use the stack of CNN to extract features at different location over the image.
The output of the CNNs at a particular layer is $Q(x;\Theta)\in \mathbb{R}^{ hw\times d}$ where $h$ and $w$ are the spatial height and width respectively, $d$ is the number of channels and $\Theta$ is the parameters of the CNN adopted.
For a specific part, we maintain a learnable vector $V_i \in \mathbb{R}^{ d\times 1}$ which is called part representation to encode the associated part.
We expect the inner-product value between $V_i$ and features in the feature map $Q(X;\Theta)$ being high in associated regions, while the value being low at other places.
As shown in Fig. ~\ref{fig.framework2}, the localization module takes image representation $Q$ and location representation $V_i$ as input and outputs the inner-product map $A_i$:
\begin{equation}
A_i(x;V_i,\Theta) = A_i(Q;V_i) = Q(x;\Theta){V_i}^T.
\label{eq:att_AV}
\end{equation}
We then normalize $A_i$ over spatial domain by the soft-max operation to derive the attention map as
\begin{equation}
A'_{i} = s(A_i).
\label{eq:att1}
\end{equation}
where $s$ is spatial Softmax function. The attention map is broadcasted over channels and the results are then multiplied back to the feature map resulting an attention weighted feature map $Q_i'$:
\begin{equation}
Q_i' =Q(x;\Theta)^T\otimes A'_{i}.
\label{eq:att2}
\end{equation}
where $\otimes$ is the operation of broadcasted multiplication. Note that, the attention weighted feature maps $Q_i'$ are different for different $i$, though the input feature map $Q$ is the same, which render output features to focus on different spatial location. 

\subsubsection{Pattern Recognition Module}
Conventional attribute recognition methods usually adopt an average pool followed by a fully-connected layer and a soft-max layer to produce final probability of an attribute.
In contrast, our method learns an attention map for each attribute to weight the feature map so that the following appearance pattern could be learned by aggregating all training data of different part locations.
Specifically, for each appearance pattern $j$ which is annotated a binary classifier label in part attribute recognition, the probability of predicted label is calculated as,
\begin{equation}
f(x;\Theta, V_i, W^T_j)= softmax(W^T_j{\bar Q_i'}).
\label{eq:att_rec}
\end{equation}
where ${\bar Q_i' \in \mathbb{R}^{ d\times 1}}$ is the global average pooling result of $Q_i'$ over spatial domain, and $W_j \in \mathbb{R}^{ d\times 2} $ are appearance specific weights for the binary classification task $j$. 

\subsection{Concept Sharing Training}
In this part, we describe the concept sharing training process.
We denote an attribute by $P_{i,j}$ where the part location index is denoted by $i$ and the pattern indexed is denoted by $j$.
The attribute recognition model has parameters $\Theta$, $V_i$, and $W_j$ to optimize.
We denote the training images for attribute $P_{i,j}$ as $X_{ij}={x_0, x_1, ..., x_{N_{ij}-1}}$ whose labels are $Y_{ij}={y_0, y_1,...,y_{N_{ij}-1}}$ where $N_{ij}$ denotes the total number of training samples for the attribute $P_{i,j}$. The $\Theta$, $V_i$, $W_j$ are trained by an end-to-end fashion using cross-entropy loss at final binary outputs of recognition modules. The overall loss is:
\begin{equation}
L =\sum_{i=0}^{N-1}L(f(x_i;\Theta, V_i, W^T_j),y_i).
\label{eq:total_loss}
\end{equation}
Where $L$ is the cross-entropy loss and $N$ is the total number of training samples satisfying: $N=\sum_{i,j}N_{ij}$.
All attributes in Eq.~\ref{eq:total_loss} share the same $\Theta$ which are used to extract the CNN features. In our concept sharing training, the localization module is learned from all training samples sharing the same part, and the pattern recognition module is learned from all training samples which sharing the same pattern. Consequently, the weights of sharing models $V_i$ and $W_j$ are updated by:
\begin{equation}
W_j^+ = W_j - \gamma\sum_{k}\sum_{n=0}^{N_{kj}-1}\frac{\partial L(f(x_n;\Theta, V_k, W^T_j),y_n) }{\partial W_j} 
\label{eq:wjnew}
\end{equation}
, and 
\begin{equation}
V_i^+ =V_i - \gamma\sum_{k}\sum_{n=0}^{N_{ik}-1}\frac{\partial L(f(x_n;\Theta, V_i, W^T_k),y_n) }{\partial Q_i'} \times \frac{\partial Q_i' }{\partial V_i}
\label{eq:vinew}
\end{equation}
It is worthwhile to mention that, if the attributes are treated to be independent as in conventional recognition frameworks, training $V_i$ and $W_j$ will only involve training samples $\left(X_{ij}, Y_{ij}\right)$. Since the number of training samples of a single attribute is usually small in practice, and therefore the performance of conventional recognition frameworks is limited by the insufficiency of training samples for each isolated attribute. In contrast, the number of training samples for a particular module would be enlarged in our conceptual  sharing framework \ie $\sum_k N_{kj}$ for the pattern recognition, and $\sum_k N_{ik}$ for part localization. The training data expansion improves the learning sufficiency with the limited data. 


\subsection{New Attribute Recognition}
In large-scale applications where the number of attributes is large, it is almost infeasible to obtain reasonable amount of qualified training data for every attribute.
In this part, we explain how CSN can be used to recognize a new attribute without requiring any training samples.
As we discussed above, the localization and recognition module are shared among different attributes, one specific combination of localization and recognition modules determines the recognition for one specific attribute, such as the attribute \textit{breast spotted} in Fig.~\ref{fig.motivation}.
Notice that we do not have any training data for \textit{breast spotted}.
However, we can still train \textit{breast} localization module and \textit{spotted} pattern module from other attributes. 
Therefore, by combining the learned location \textit{forehead} and pattern \textit{spotted}, \textit{forehead spotted} could be understandably recognized without any such training data. 
Specifically, attribute $P_{i,j}$ recognition model is determined by its parameters $\Theta$, $W_j$ and $V_i$. Since $W_j$ could be learned from attributes $P_{\alpha,j}$($\alpha \neq
i$), and $V_i$ could be learned from attributes  $P_{i,\beta}$($\beta \neq
j$), attribute $P_{i,j}$ recognition model could be determined even without any $P_{i,j}$ data.

\section{Experiments}
We conduct experiments on three attribute recognition datasets including CUB-200-2011~\cite{wah2011caltech}, CelebA~\cite{liu2015deep} and our new SurveilA dataset.
Since the positive and negative samples are highly imbalanced in CUB-200-2011 and SurveilA, we use average precision as our main evaluation metric.

\subsection{Datasets}
The CUB-200-2011 dataset~\cite{wah2011caltech} contains 11,788 images of 200 bird categories, where 5994 images are selected for training and the rest 5794 images for testing.
Each image is annotated with 312 attributes among which 278 are part related attributes.
These part related attributes are all binary attributes indicating whether a specific appearance pattern exists in one particular part such as attribute \textit{wing-black} tells whether the wing is black.
During experiments,we observe that the back and tail attributes are nosily labelled.
Therefore, we exclude these labels and conduct experiments on the remaining 204 attributes.
This is the main dataset for our experimental comparison and ablation study.

CelebA~\cite{liu2015deep} consists of 202,599 face images and 40 binary face attributes. 16,000 images are selected for training and 20,000 images for testing and validation. 

Our new SurveilA dataset contains 75,000 images with 10 binary attributes, among which 70,000 are selected for training and the rest 5,000 for testing.
The dataset focuses on human part attribute (\eg whether carrying an item, whether wearing shorts or pants), collected under real surveillance scenarios with large pose and appearance variations. 
We will release the dataset to facilitate the research of attribute recognition.

\subsection{Implementation Details}
 We first resize the image to be $512 \times 512$, then randomly cropped $446 \times 446$ for training. 
 We use ResNext50~\cite{xie2017aggregated} as the visual representation module for feature extraction.
 The outputs from the layer `conv5' in ResNext50 are used as the feature representation for part localization \ie $Q$ in eq.~\ref{eq:att_AV}.
 The network is trained for 100 epochs with ADAM\cite{kingma2014adam} where initial learning rate is set to 0.0001 with a learning rate decay of 0.1 after 50 epochs.
 
\subsection{Experiments on CUB-200-2011 Dataset}



\subsubsection{Study the Number of Training Samples}

In this part, we empirically study the the effectiveness of CSN as the number of training samples varies.
In order to study this, we select attributes \textit{belly-solid}, \textit{breast-white}, \textit{bill-grey} and \textit{bill-black} which could share the same part and have relatively larger positive samples.
All experiments are run on joint training of 91 attributes(\ie all bill relevant attributes, all grey attributes, all black attributes, all belly, all breast, all white).
We evaluate the performance of the CSN w/o(without) share, CSN w/part share, CSN w/part + pattern share on the 3 attributes selected.
The results are shown in Tab.~\ref{tab:ablation200} and Tab.~\ref{tab:ablation500}
We see the benefits of sharing attributes is more obvious when the number of training samples is small.
The pattern features extracted at different location still varies, forcing them to be the same pattern when they have relatively large data will harm the performance.
But when the training data size decreases, their own data is insufficient to learn the pattern module.
Pattern sharing will improve the learning efficiency of the pattern module. 

We also try to decrease the training samples to zero. CSN still obtains promising results for \textit{belly-solid} (84.7\% AP), \textit{breast-white} (79.9\% AP), \textit{bill-grey} (46.3\% AP), \textit{bill-black} (64.2\% AP) that are comparable to the supervised training.
We will further investigate the effectiveness of CSN for zero-shot recognition in the following section.

\begin{table}
    \centering
    \rowcolors{2}{}{yelloworange!25}
    \addtolength{\tabcolsep}{-2pt}
    \begin{tabular}{l c c c}
        \toprule[0.2 em]
        attribute & no share & part share & pattern+part share \\
        \toprule[0.2 em]
        belly-solid & 83.6\% & 85.9\% & 85.5\% \\
        breast-white & 81.3\% & 82.1\% & 81.9\% \\
        bill-grey & 44.7\%  & 46.0\% & 47.0\% \\
        bill-black & 78.2\% & 79.4\% & 77.0\% \\
        \bottomrule[0.1 em]
    \end{tabular}
    \vspace{1pt}
    \caption{Average precision given 500 training images}
    \vspace{5pt}
	\label{tab:ablation500}
    \begin{tabular}{l c c c}
        \toprule[0.2 em]
        attribute & no share & part share & pattern+part share \\
        \toprule[0.2 em]
        belly-solid & 79.2\% & 84.0\% & 84.9\%\\
        breast-white & 76.5\% & 79.7\% & 80.5\% \\
        bill-grey & 38.5\%  & 42.2\%  & 46.8\% \\
        bill-black & 74.9\% & 78.6\%  & 76.0\% \\
        \bottomrule[0.1 em]
    \end{tabular}
    \vspace{1pt}
    \caption{Average precision given 200 training images}
	\label{tab:ablation200}
\end{table}

\subsubsection{Study the Number of Sharing Attributes}
In this part, we empirically study the effects of the number of sharing attributes on the overall performance. 
In Tab.~\ref{tab:number}, we compare sharing different number of attributes in pattern module and localization module. We select attribute \textit{bill-curved},\textit{bill-brown},\textit{bill-orange} and \textit{bill-red}. We perform experiments on joint training of all bill attributes, all brown attributes, all orange attributes and all red attributes. We compare the four attributes performance under different number of sharing part attributes and sharing pattern attributes. 

From (a) - (c) in Tab.~\ref{tab:number}, as the number of sharing part attributes increases, the overall performance steadily increases. This illustrates that it is more effective when number of sharing part attributes become larger. Note that as the number of sharing pattern attributes increases to be 9, the improvement is observed, this also indicates the sharing pattern is effective. 

\begin{figure}
	\centering
	\includegraphics[width=0.75\linewidth]{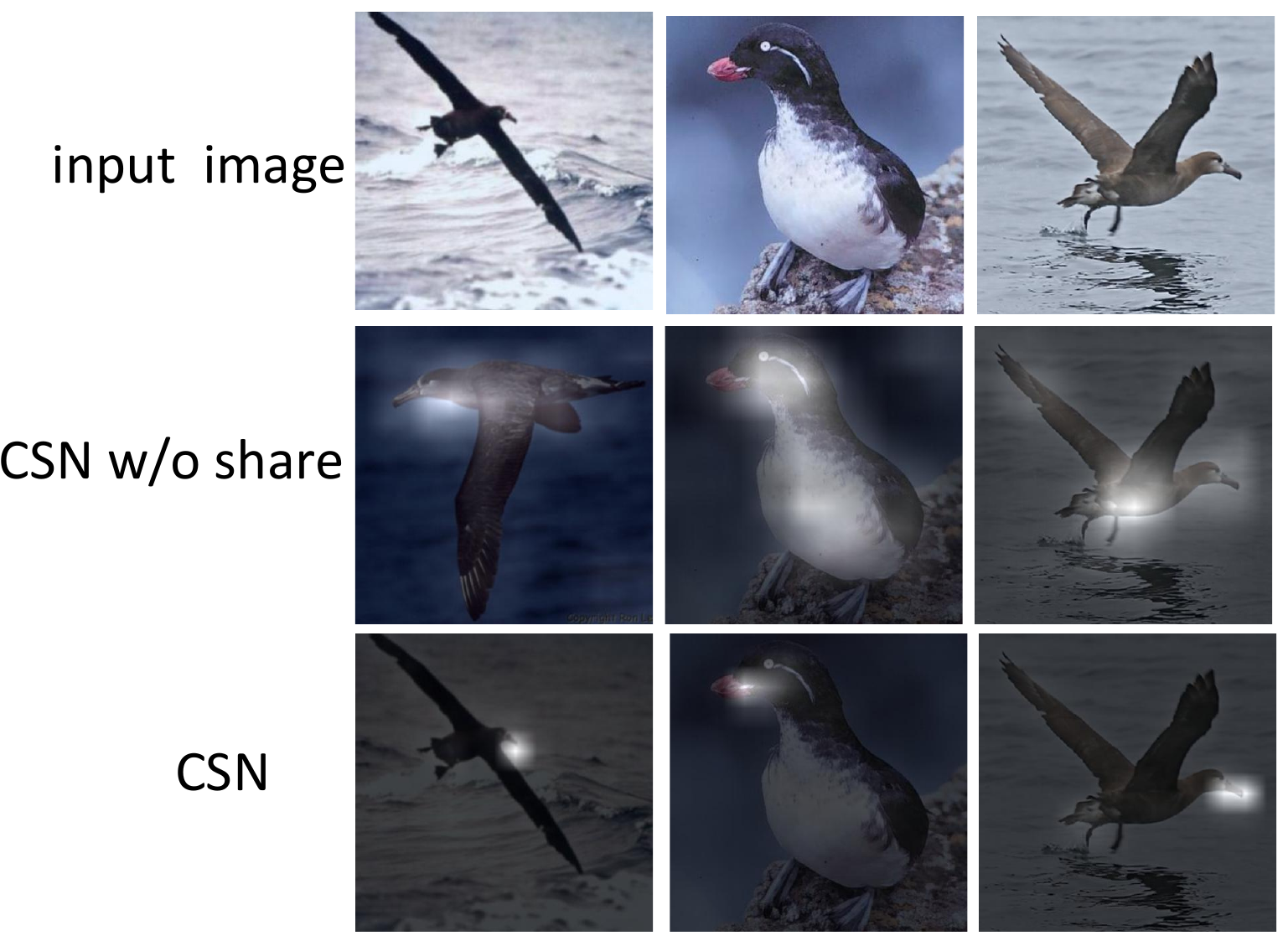}
	\caption{Comparison of heat map generated by CSN w/o sharing and full CSN for the part ``bill".}
	\label{fig.att}
\end{figure}


\begin{table}
    \centering
    \small
    \rowcolors{2}{}{yelloworange!25}
    \addtolength{\tabcolsep}{-4.5pt}
    \begin{tabular}{c c c c c c}
        \toprule[0.2 em]
        Baseline & CAM~\cite{zhou2016learning} & STN~\cite{jaderberg2015spatial} & PANDA~\cite{zhang2014panda} & R-CNN~\cite{zhang2014part} & CSN \\
        \toprule[0.2 em]
        63.1\% & 63.2\% & 63.7\% & 64.9\% & 65.5\%* & 65.2\% \\
        \bottomrule[0.1 em]
    \end{tabular}
    \vspace{1pt}
    \caption{Average Precision(AP) comparison.
    The numbers are shown only for the 32 attributes that contain more than 1000 positive samples in this dataset.
    If we perform such comparison for attributes with smaller training set, the baseline always produces very poor result.}
	\label{tab:32}
\end{table}

\begin{table}
    \centering
    \rowcolors{2}{}{yelloworange!25}
    \addtolength{\tabcolsep}{0pt}
    \begin{tabular}{c c c c}
        \toprule[0.2 em]
        CSN w/o share & CSN & CSN-soft & CSN-soft-1 \\
        \toprule[0.2 em]
         63.9\% & 65.2\% & 65.1\% & 63.8\% \\
        \bottomrule[0.1 em]
    \end{tabular}
    \vspace{1pt}
    \caption{Average Precision(AP) comparison.
    The numbers are shown only for the 32 attributes that contain more than 1000 positive samples in this dataset.
    If we perform such comparison for attributes with smaller training set, the baseline always produces very poor result.}
	\label{tab:33}
\end{table}

\subsubsection{Comparison with the State-of-the-art}

\begin{figure*}
	\centering
	\includegraphics[width=13.9cm]{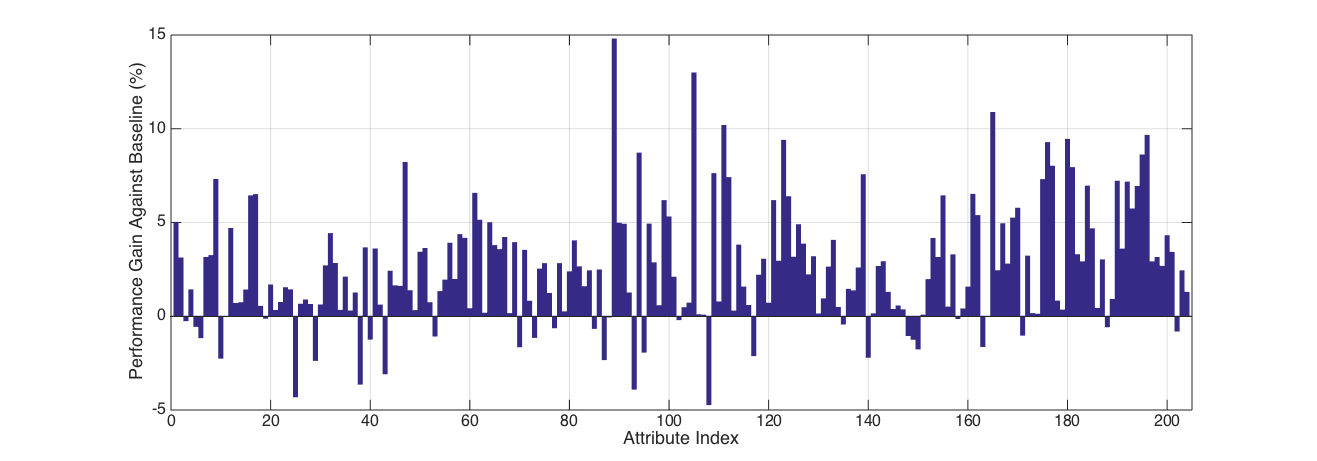}
	\caption{CSN performance gain against baseline on joint training of 204 attributes. The baseline method refers to a multi task learning framework with 204 branches. 
	Please refer to the supplementary materials for the specific name of each attribute id.}
	\label{fig.204}
	\vspace{-1mm}
\end{figure*}

We first compare with state-of-the-art recognition method with only image level annotation CAM~\cite{zhou2016learning}, Spatial Transform Network(STN)~\cite{jaderberg2015spatial} and our baseline (a multi task learning framework with different branches).
STN aims at explicitly localizing important regions for recognition. We used the public implementation and replace the recognition by our part attribute recognition. 
Our CAM implementation follows~\cite{zhou2016learning} (i.e.
localize and crop) for recognition. 
Tab.~\ref{tab:32} shows the performance of our method against other methods.
Since the baseline always shows very low performance with attributes with small number of training images,
we select the attributes with more than 1000 positive images for a fair comparison.
Such selection leads to 32 attributes (e.g. as white relevant attributes and black relevant attributes) in the experiments.
CSN produces AP of 65.2\% compared to the competitors: 63.2\% from CAM, 63.7\% from STN and 63.1\% of CSN baseline. This illustrates the advantage of CSN over alternatives given a reasonably large training set. We investigate all 204 part attributes to further understand the overall performance of different methods.
Fig.~\ref{fig.204} shows the AP difference between of CSN and the baseline on all 204 attributes. Significant improvement over the baseline is observed on average. 

We then compare with state-of-the-art attributes recognition methods~\cite{zhang2014panda,zhang2014part}. We used the public implementation~\cite{zhang2014panda,bourdev2009poselets,zhang2014part} to train a parts detector(i.e. poselets~\cite{bourdev2009poselets} and R-CNN~\cite{zhang2014part}
) with part annotation and then recognize the attributes. Without part annotation, we still obtain comparable performance with PANDA~\cite{zhang2014panda} and R-CNN based method ~\cite{zhang2014part}. Since they both depend on the part annotations to train the part detector, they fail when the part annotation is not available. Furthermore, they both can not be used for zero shot recognition while ours can. 

We also visualize the localization heat map(\ie the attention map) obtained by CSN w/o share and CSN in Fig. ~\ref{fig.att}.
The localization heat map obtained by CSN is obviously better than that obtained by CSN w/o share. This further verifies the effectiveness of the concept sharing. 

\begin{table}
    \small
	\centering
	\rowcolors{2}{}{yelloworange!25}
    \addtolength{\tabcolsep}{-4.5pt}
    \begin{tabular}{l c c c c}
    \toprule[0.2 em]
    & bill-curved & bill-brown & bill-orange & bill-red\\
    \toprule[0.2 em]
    (a) $N_1=0$, $N_2=0$ & 19.1\% & 25.6\% &38.6\% & 35.8\% \\
    (b) $N_1=4$, $N_2=0$ & 20.1\% & 25.9\% & 39.4\% & 36.7\% \\
    (c) $N_1=24$, $N_2=0$ &21.4\% & 26.7\% & 42.7\% & 38.1\% \\
    (d) $N_1=24$, $N_2=9$ & - & 26.9\% & 43.0\% & 38.3\% \\
    \bottomrule[0.1 em]
    \end{tabular}
    \vspace{1pt}
    \caption{Average Precision(AP) comparison with different sharing attribute number. $N_1$ is the number of sharing part attributes, and $N_2$ is the number of sharing pattern attributes. Note: \textit{curved} attribute only exists in \textit{bill-curved}, so (d) for \textit{bill-curved} is not available.}
	\label{tab:number}
\end{table}

\begin{figure*}
	\centering
	\includegraphics[width=13.9cm]{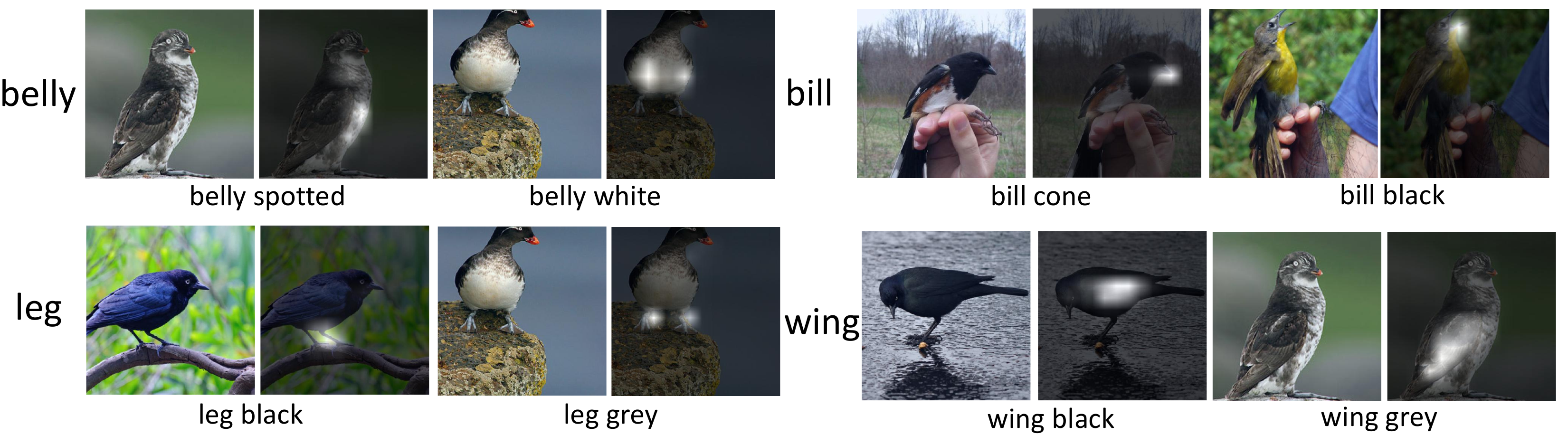}
	\caption{Qualitative results of CSN on CUB-200-2011 test set.
	For each pair of images in the figure, the left shows the input image.
	The right shows the location heat map predicted by CSN.
	The bottom text shows the predicted attribute. More visualization results are in supplementary materials.}
	\label{fig:cub}
\end{figure*}
\begin{table*}
	\centering
    \rowcolors{2}{}{yelloworange!25}
    \addtolength{\tabcolsep}{0pt}
    \begin{tabular}{c c c c c c c c c c c}
    \toprule[0.2 em]
    Attribute ID & 1 & 2 & 3 & 4 & 5 & 6 & 7 & 8 & 9 & 10     \\
    \toprule[0.2 em]
    \# of training imgs & 51968 & 12050 & 4982 & 3183 & 1334 & 542 & 320 & 276 & 209 & 147 \\
    Baseline & 89.5\% & 47.1\% & 57.0\% & 38.0\% & 49.0\% & 7.1\% & 5.9\% & 6.3\% & 2.2\% & 0.5\% \\
    CSN & {\bf 96.3\%} & {\bf 65.8\%} & {\bf 75.6\%} & {\bf 67.3\%} & {\bf 84.1\%} & {\bf 23.4\%} & {\bf 32.2\%} & {\bf 38.3\%} & {\bf 7.9\%} & {\bf 30.3\%} \\
    \bottomrule[0.1 em]
    \end{tabular}
    \caption{Average Precision(AP) on our new human attribute test set. The attributes 1-10 are some of the most useful ones in security surveillance applications:
    1~\textit{the length of sleeve (short / long)},
    2~\textit{the length of pants (long / short)},
    3~\textit{using a cell phone or not},
    4~\textit{carrying an item or not},
    5~\textit{dragging a luggage or not},
    6~\textit{smoking or not},
    7~\textit{wearing a glove or not},
    8~\textit{holding a baby in arm or not},
    9~\textit{wearing a mask or not},
    10~\textit{holding an umbrella or not}.
    The 2nd row shows the number of the training samples.
    It can be observed that CSN obtains higher performance gain for attributes with fewer training samples.}
	\label{tab:human}
\end{table*}
\begin{table}
    \small
    \centering
    \rowcolors{2}{}{yelloworange!25}
    \addtolength{\tabcolsep}{-3pt}
    \begin{tabular}{c c c c c}
        \toprule[0.2 em]
        Baseline & Multi-task~\cite{hand2017attributes} & Fully~\cite{lu2017fully} & CSN w/o share & CSN\\
        \toprule[0.2 em]
        91.1\% & 91.2\% & 91.3\% & 91.7\% & \bf{91.8}\%\\
        \bottomrule[0.1 em]
    \end{tabular}
    \vspace{1pt}
    \caption{Comparison of attribute classification accuracy on CelebA test set. }
    \label{tab:celeb}
\end{table}
\begin{table}
	\centering
	\rowcolors{2}{}{yelloworange!25}
    \addtolength{\tabcolsep}{-4.5pt}
    \begin{tabular}{c c c c}
    \toprule[0.2 em]
    attribute & \#images & zero-shot & supervised \\
    \toprule[0.2 em]
    belly\_solid          & 2455       & 86.1\%    & 88.1\%     \\
    upperparts\_black     & 1561       & 74.5\%    & 78.8\%     \\
    crown\_black          & 1434       & 75.7\%    & 82.9\%     \\
    underparts\_black     & 730        & 67.3\%    & 74.5\%     \\
    breast\_multi-colored & 626        & 30.8\%    & 37.9\%     \\
    crown\_buff           & 413        & 38.1\%    & 39.2\%     \\
    underparts\_brown     & 360        & 41.2\%    & 47.8\%     \\
    belly\_striped        & 319        & 36.5\%    & 41.5\%     \\
    wing\_yellow          & 316        & 52.3\%    & 63.9\%     \\
    belly\_brown          & 313        & 36.7\%    & 44.8\%     \\
    wing\_spotted         & 300        & 37.0\%    & 51.6\%     \\
    bill\_yellow          & 215        & 9.2\%     & 50.3\%     \\
    throat\_red           & 175        & 65.6\%    & 70.0\%     \\
    belly\_red            & 140        & 68.3\%    & 73.1\%     \\
    wing\_olive           & 127        & 33.2\%    & 30.6\%     \\
    upperparts\_orange    & 81         & 18.7\%    & 28.1\%     \\
    wing\_iridescent      & 74         & 12.0\%    & 13.0\%     \\
    belly\_olive          & 67         & 13.9\%    & 23.9\%     \\
    underparts\_green     & 38         & 25.2\%    & 16.0\%     \\
    forehead\_purple      & 22         & 11.2\%    & 4.2\%      \\
    \bottomrule[0.1 em]
    \end{tabular}
    \caption{Average Precision(AP) comparison between zero shot learning and supervised learning on 20 attributes. Zero shot is CSN with part and pattern sharing, supervised is CSN w/o sharing.}
	\label{tab:zero}
\end{table}

\subsubsection{Soft Sharing among Attributes}
In the above section, attributes with the same part are manually grouped to the same part localization module, and attributes with the same pattern are manually grouped to the same pattern module. Accordingly, attributes with different concepts (localization / pattern) can not share the same module. 
In this part, we study the soft sharing among parts. We add a learnable soft weight vector $T_i = t_1, t_2, ..,t_m$ for attribute $k$, where $m$ is number of parts modules. This weight vector is used to combine the attention map from different part. That is 

\begin{equation}
A_k' =\sum_{i=0}^{m-1}t_iA_i.
\label{eq:combine}
\end{equation}

We first initialize the vector to be one hot vector(\ie one value is 1, others are 0 in $T_i$). Let the attributes sharing the same part have the same initialization, and let them learnable during the training. This obtains 65.1\%(\ie CSN-soft in Tab.~\ref{tab:32}) which is comparable with CSN. We then initialize the vectors as 1(\ie all values in $T_i$ are initialized as 1). This indicates that we do not have the prior knowledge which attributes are of the same part.  We obtain 63.8\%(\ie CSN-soft-1 in Tab.~\ref{tab:32}) which is lower than 65.1\% of the previous one. This indicates that this prior knowledge provide important information. 

\subsubsection{Zero-Shot Attribute Recognition}
In this part, we study the capability of CSN on zero shot attribute recognition.
The experiments are carried out on all 204 attributes where
we randomly select 20 attributes as unseen attributes and leave the other 184 attributes for training the network.
In Tab.~\ref{tab:zero}, zero-shot algorithm refers to CSN w/ sharing pattern and part with no training data on the 20 attributes, this is the algorithm for zero shot in Tab.~\ref{tab:zero}, and supervised refers to CSN w/o sharing trained on all available data.
Since the performance gap between zero-shot and supervised is highly relevant to the size of train data, we also list the number of positive samples in statistics.
As shown in Tab.~\ref{tab:zero}, zero-shot obtains promising results on the condition of zero shot learning.
Note that attributes with very small data, such as \textit{'forehead purple'},  \textit{'wing olive'}, \textit{'underparts green'}, zero shot even outperforms results trained on their own data.
For most of attributes, zero shot algorithm is surprisingly effective as it shows accuracy comparable with trained on their own data.

\begin{figure}
	\centering
	\includegraphics[scale=0.4]{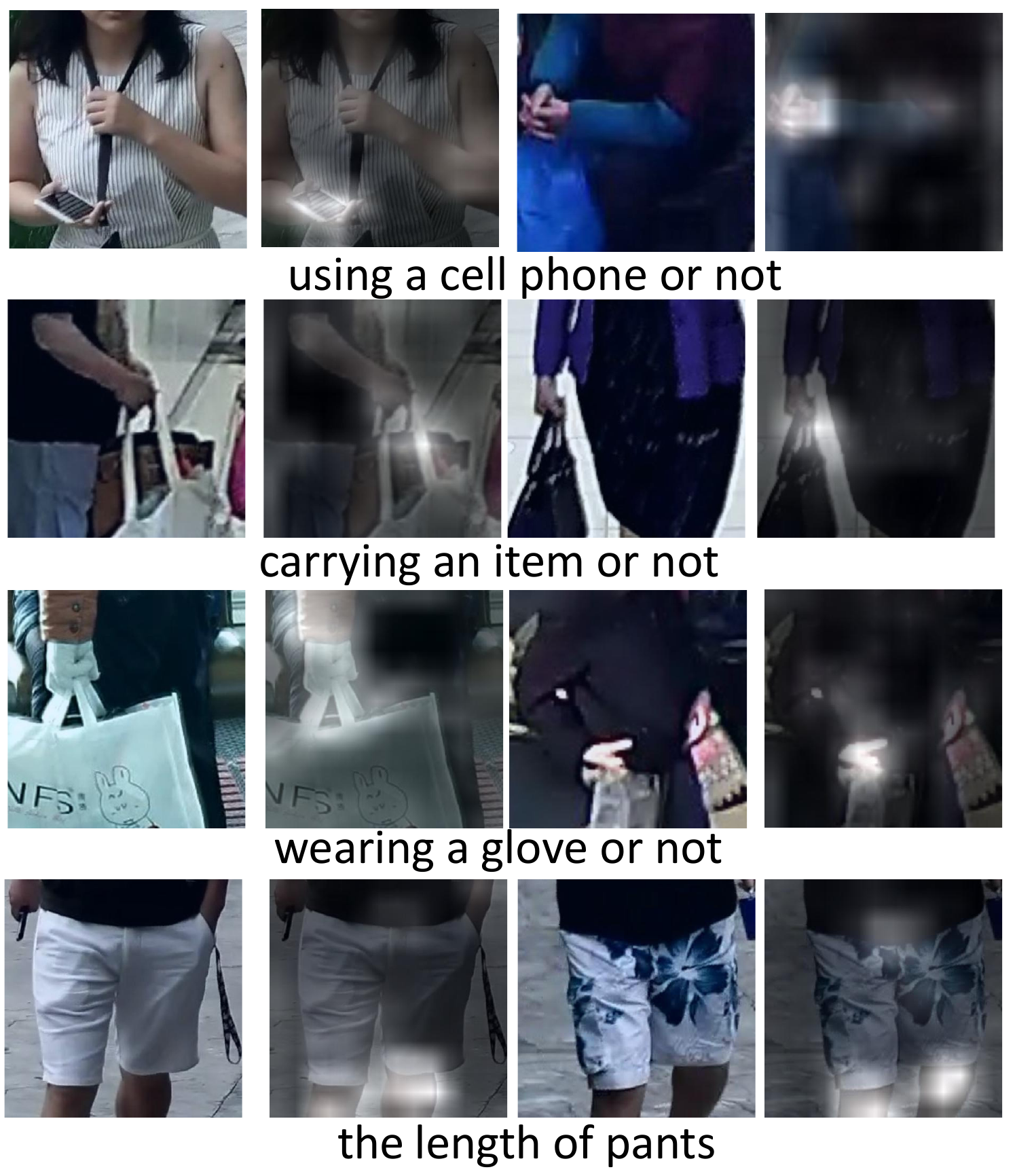}
	\caption{Exemplar images of our human dataset. The heat map placed on the right of each image visualizes the localization information predicted in the CSN inference process.
	We see that most attributes are only associated with a very small area in images.}
	\label{fig:SurveilA}
	\vspace{-3mm}
\end{figure}

	

 \subsection{Experiments on CelebA}

Our method can also be applied on the general attributes(i.e. global and parts attributes), so we also perform experiments on a general attributes dataset CelebA~\cite{liu2015deep}. We follow the protocol in~\cite{hand2017attributes}. In table~\ref{tab:celeb}, we evaluate the average accuracy which is usually reported for CelebA. Our CSN obtains better performance than our baseline and beats the state of the art methods. This is because all other methods fail to explicit localize the important regions for recognition. 
In CSN, we group the parts attributes to share the same localization module such as nose related and mouth related. We observe further improvement by sharing the localization module. This indicates that the concept sharing is still effective. 

\subsection{Experiments on SurveilA Dataset}

Tab.~\ref{tab:human} shows that in this dataset CSN achieves 51.2\% mAP compared to baseline mAP 30.3\%.
Such a huge performance improvement is achieved since most of these human attributes only depend on a small part of the image (e.g. \textit{wearing a mask or not} are only related to the face area). Hence, the localization function of CSN establishes higher importance than the CUB-200-2011 dataset.  
In Fig.~\ref{fig:SurveilA}, we visualize the parts localization obtained by our methods. The experiments on humans further verify that our proposed method is effective and reliable for recognizing parts of different types of objects.  

	\vspace{-3mm}

\section{Conclusion}
In this paper, we proposed a new neural network structure (CSN) for part attribute recognition.
By identifying part locations and appearance patterns from the training data that do not explicitly label these two concepts, CSN can increase part attribute recognition accuracy especially if the number of labels is small. 
In the special case of data limitation where none data is available, CSN is still valid to recognize attributes (\ie zero-shot part attribute recognition).
\subsubsection*{Acknowledgements}

This work was supported in part by National Science 
Foundation grant IIS-1619078, IIS-1815561, and the Army
Research Office ARO W911NF-16-1-0138.

{\small
\bibliographystyle{ieee_fullname}
\bibliography{egbib}
}
\onecolumn

\appendix

\section{Experimental Attributes}
In Figure 5, we mentioned that we performed experiments on 204 attributes. The 1-204 attributes are: \textit{1 curved\_(up\_or\_down)} , \textit{2 bill\_dagger} , \textit{3 bill\_hooked} , \textit{4 bill\_needle} , \textit{5 bill\_hooked\_seabird} , \textit{6 bill\_spatulate} , \textit{7 bill\_all-purpose} , \textit{8 bill\_cone} , \textit{9 bill\_specialized} , \textit{10 wing\_blue} , \textit{11 wing\_brown} , \textit{12 wing\_iridescent} , \textit{13 wing\_purple} , \textit{14 wing\_rufous} , \textit{15 wing\_grey} , \textit{16 wing\_yellow} , \textit{17 wing\_olive} , \textit{18 wing\_green} , \textit{19 wing\_pink} , \textit{20 wing\_orange} , \textit{21 wing\_black} , \textit{22 wing\_white} , \textit{23 wing\_red} , \textit{24 wing\_buff} , \textit{25 upperparts\_blue} , \textit{26 upperparts\_brown} , \textit{ 27 upperparts\_iridescent} , \textit{28 upperparts\_purple} , \textit{29 upperparts\_rufous} , \textit{30 upperparts\_grey} , \textit{31 upperparts\_yellow} , \textit{32 upperparts\_olive} , \textit{33 upperparts\_green} , \textit{34 upperparts\_pink} , \textit{35 upperparts\_orange} , \textit{36 upperparts\_black} , \textit{37 upperparts\_white} , \textit{38 upperparts\_red} , \textit{39 upperparts\_buff} , \textit{40 underparts\_blue} , \textit{41 underparts\_brown} , \textit{42 underparts\_iridescent} , \textit{43 underparts\_purple} , \textit{44 underparts\_rufous} , \textit{45 underparts\_grey} , \textit{46 underparts\_yellow} , \textit{47 underparts\_olive} , \textit{48 underparts\_green} , \textit{49 underparts\_pink} , \textit{50 underparts\_orange} , \textit{51 underparts\_black} , \textit{52 underparts\_white} , \textit{53 underparts\_red} , \textit{54 underparts\_buff} , \textit{55 breast\_solid} , \textit{56 breast\_spotted} , \textit{57 breast\_striped} , \textit{58 breast\_multi-colored} , \textit{59 head\_spotted} , \textit{60 head\_malar} , \textit{61 head\_crested} , \textit{62 head\_masked} , \textit{63 head\_unique\_pattern} , \textit{64 head\_eyebrow} , \textit{65 head\_eyering} , \textit{66 head\_plain} , \textit{67 head\_eyeline} , \textit{68 head\_striped} , \textit{69 head\_capped} , \textit{70 breast\_blue} , \textit{71 breast\_brown} , \textit{72 breast\_iridescent} , \textit{73 breast\_purple} , \textit{74 breast\_rufous} , \textit{75 breast\_grey} , \textit{76 breast\_yellow} , \textit{77 breast\_olive} , \textit{78 breast\_green} , \textit{79 breast\_pink} , \textit{80 breast\_orange} , \textit{81 breast\_black} , \textit{82 breast\_white} , \textit{83 breast\_red} , \textit{84 breast\_buff} , \textit{85 throat\_blue} , \textit{86 throat\_brown} , \textit{87 throat\_iridescent} , \textit{88 throat\_purple} , \textit{89 throat\_rufous} , \textit{90 throat\_grey} , \textit{91 throat\_yellow} , \textit{92 throat\_olive} , \textit{93 throat\_green} , \textit{94 throat\_pink} , \textit{95 throat\_orange} , \textit{96 throat\_black} , \textit{97 throat\_white} , \textit{98 throat\_red} , \textit{99 throat\_buff} , \textit{100 eye\_blue} , \textit{101 eye\_brown} , \textit{102 eye\_purple} , \textit{103 eye\_rufous} , \textit{104 eye\_grey} , \textit{105 eye\_yellow} , \textit{106 eye\_olive} , \textit{107 eye\_green} , \textit{108 eye\_pink} , \textit{109 eye\_orange} , \textit{110 eye\_black} , \textit{111 eye\_white} , \textit{112 eye\_red} , \textit{113 eye\_buff} , \textit{ 114 bill\_length::about\_the\_same\_as\_head} , \textit{115 bill\_length::longer\_than\_head} , \textit{116 bill\_length::shorter\_than\_head} , \textit{117 forehead\_color::blue} , \textit{118 forehead\_color::brown} , \textit{119 forehead\_color::iridescent} , \textit{120 forehead\_purple} , \textit{121 forehead\_rufous} , \textit{122 forehead\_grey} , \textit{123 forehead\_yellow} , \textit{124 forehead\_olive} , \textit{125 forehead\_green} , \textit{126 forehead\_pink} , \textit{127 forehead\_orange} , \textit{128 forehead\_black} , \textit{129 forehead\_white} , \textit{130 forehead\_red} , \textit{131 forehead\_buff} , \textit{132 belly\_blue} , \textit{133 belly\_brown} , \textit{134 belly\_iridescent} , \textit{135 belly\_purple} , \textit{136 belly\_rufous} , \textit{137 belly\_grey} , \textit{138 belly\_yellow} , \textit{139 belly\_olive} , \textit{140 belly\_green} , \textit{141 belly\_pink} , \textit{142 belly\_orange} , \textit{143 belly\_black} , \textit{144 belly\_white} , \textit{145 belly\_red} , \textit{146 belly\_buff} , \textit{147 wing\_rounded-wings} , \textit{148 wing\_pointed-wings} , \textit{149 wing\_broad-wings} , \textit{150 wing\_tapered-wings} , \textit{151 wing\_shape::long-wings} , \textit{152 belly\_solid} , \textit{153 belly\_spotted} , \textit{154 belly\_striped} , \textit{155 belly\_multi-colored} , \textit{156 leg\_blue} , \textit{157 leg\_brown} , \textit{158 leg\_iridescent} , \textit{159 leg\_purple} , \textit{160 leg\_rufous} , \textit{161 leg\_grey} , \textit{162 leg\_yellow} , \textit{163 leg\_olive} , \textit{164 leg\_green} , \textit{165 leg\_pink} , \textit{166 leg\_orange} , \textit{167 leg\_black} , \textit{168 leg\_white} , \textit{169 leg\_red} , \textit{170 leg\_buff} , \textit{171 bill\_blue} , \textit{172 bill\_brown} , \textit{173 bill\_iridescent} , \textit{174 bill\_purple} , \textit{175 bill\_rufous} , \textit{176 bill\_grey} , \textit{177 bill\_yellow} , \textit{178 bill\_olive} , \textit{179 bill\_green} , \textit{180 bill\_pink} , \textit{181 bill\_orange} , \textit{182 bill\_black} , \textit{183 bill\_white} , \textit{184 bill\_red} , \textit{185 bill\_buff} , \textit{186 crown\_blue} , \textit{187 crown\_brown} , \textit{188 crown\_iridescent} , \textit{189 crown\_purple} , \textit{190 crown\_rufous} , \textit{191 crown\_grey} , \textit{192 crown\_yellow} , \textit{193 crown\_olive} , \textit{194 crown\_green} , \textit{195 crown\_pink} , \textit{196 crown\_orange} , \textit{197 crown\_black} , \textit{198 crown\_white} , \textit{199 crown\_red} , \textit{200 crown\_buff} , \textit{201 wing\_solid} , \textit{202 wing\_spotted} , \textit{203 wing\_striped} , \textit{204 wing\_multi-colored}. \\
\section{Exemplar images}
Following are some exemplar images of our human dataset and CUB bird dataset. The heat map placed on the right of each image visualizes the localization information predicted in the CSN inference process. The heat map is $A_i'$ in line 299 and Figure 3. 
\begin{figure}
	\centering
	\includegraphics[scale=0.45]{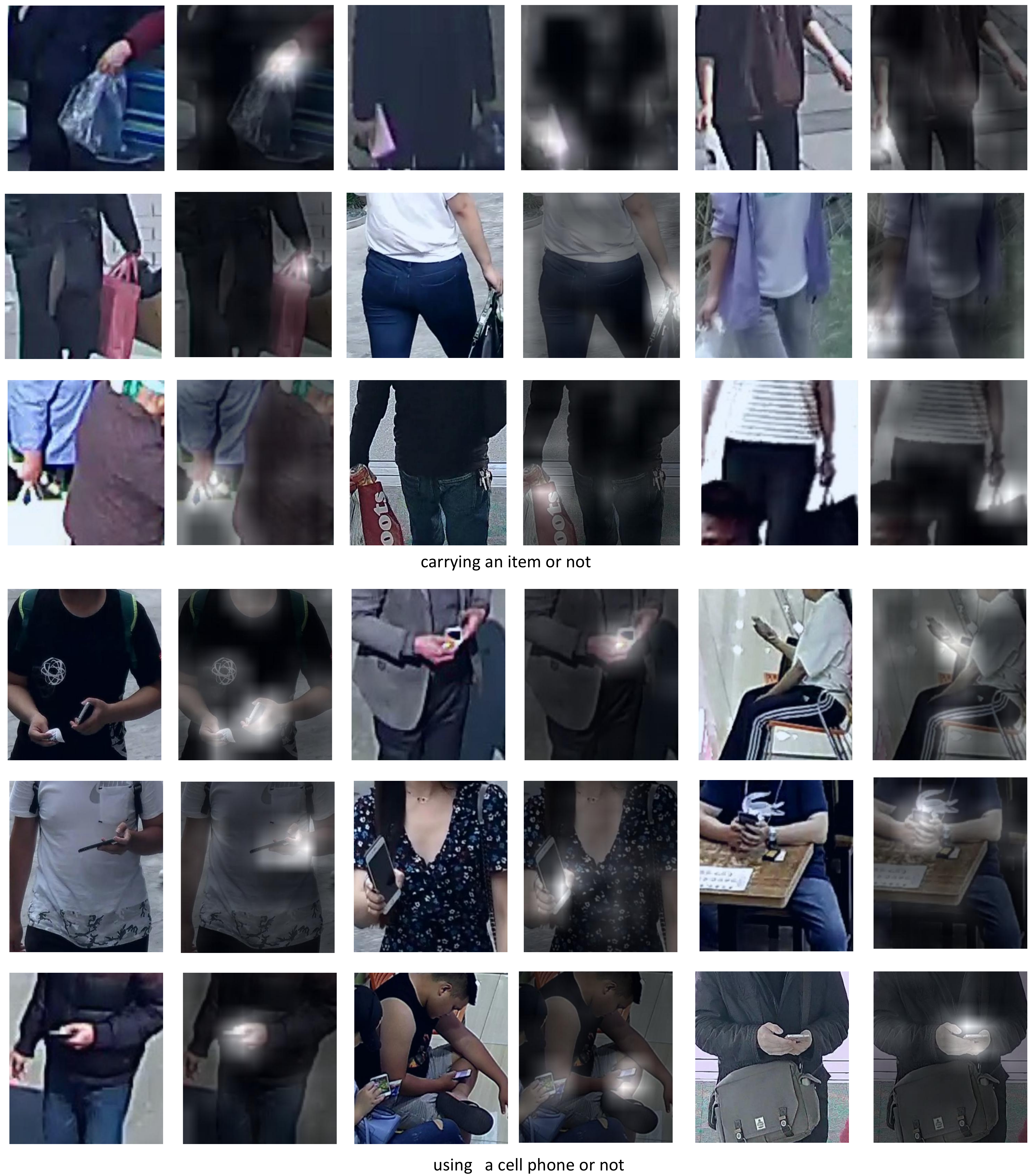}
	\caption{Exemplar images of our human dataset.  }
	\vspace{-3mm}
\end{figure}

\begin{figure}
	\centering
	\includegraphics[scale=0.45]{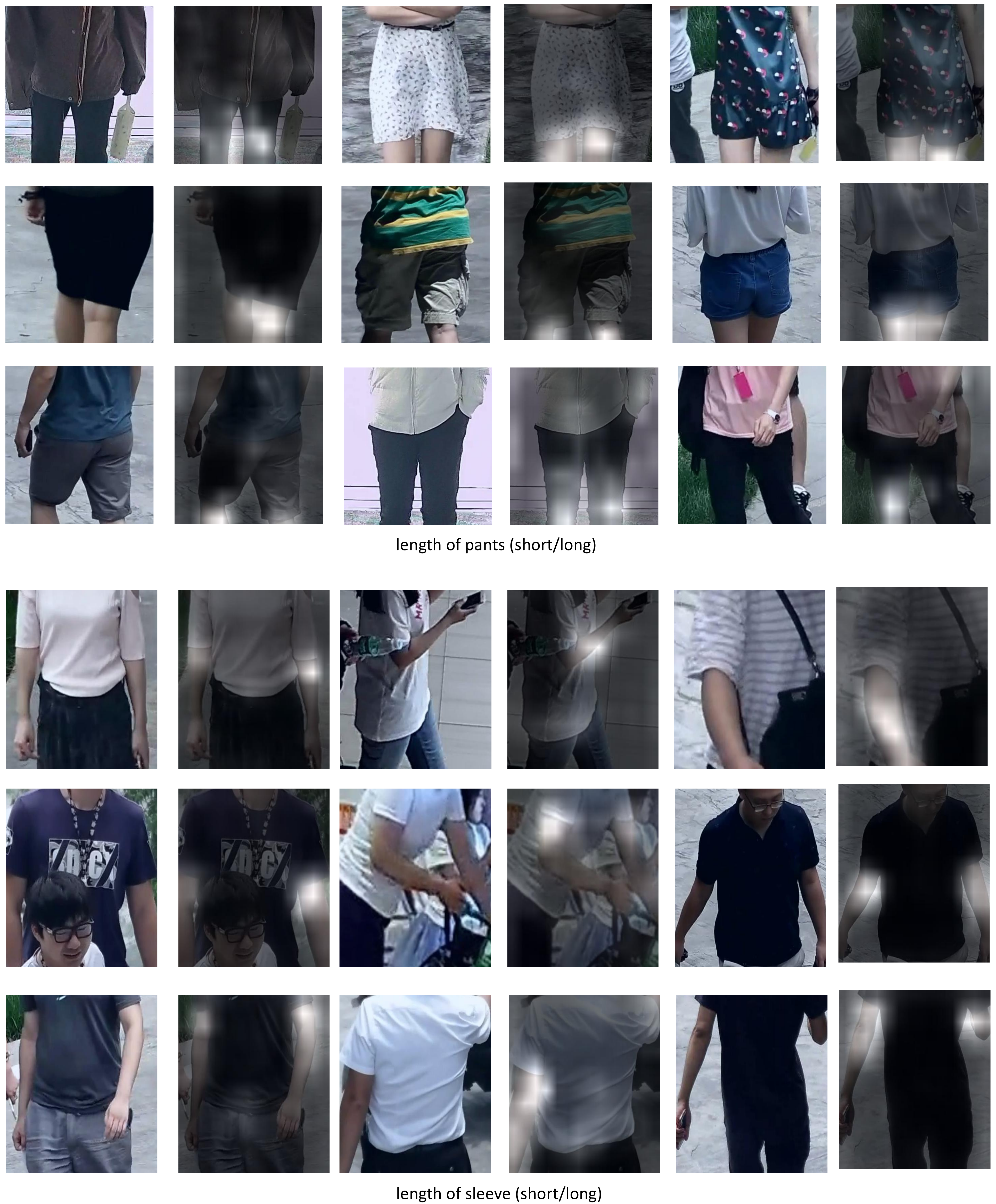}
	\caption{Exemplar images of our human dataset.}
	\vspace{-3mm}
\end{figure}

\begin{figure}
	\centering
	\includegraphics[scale=0.45]{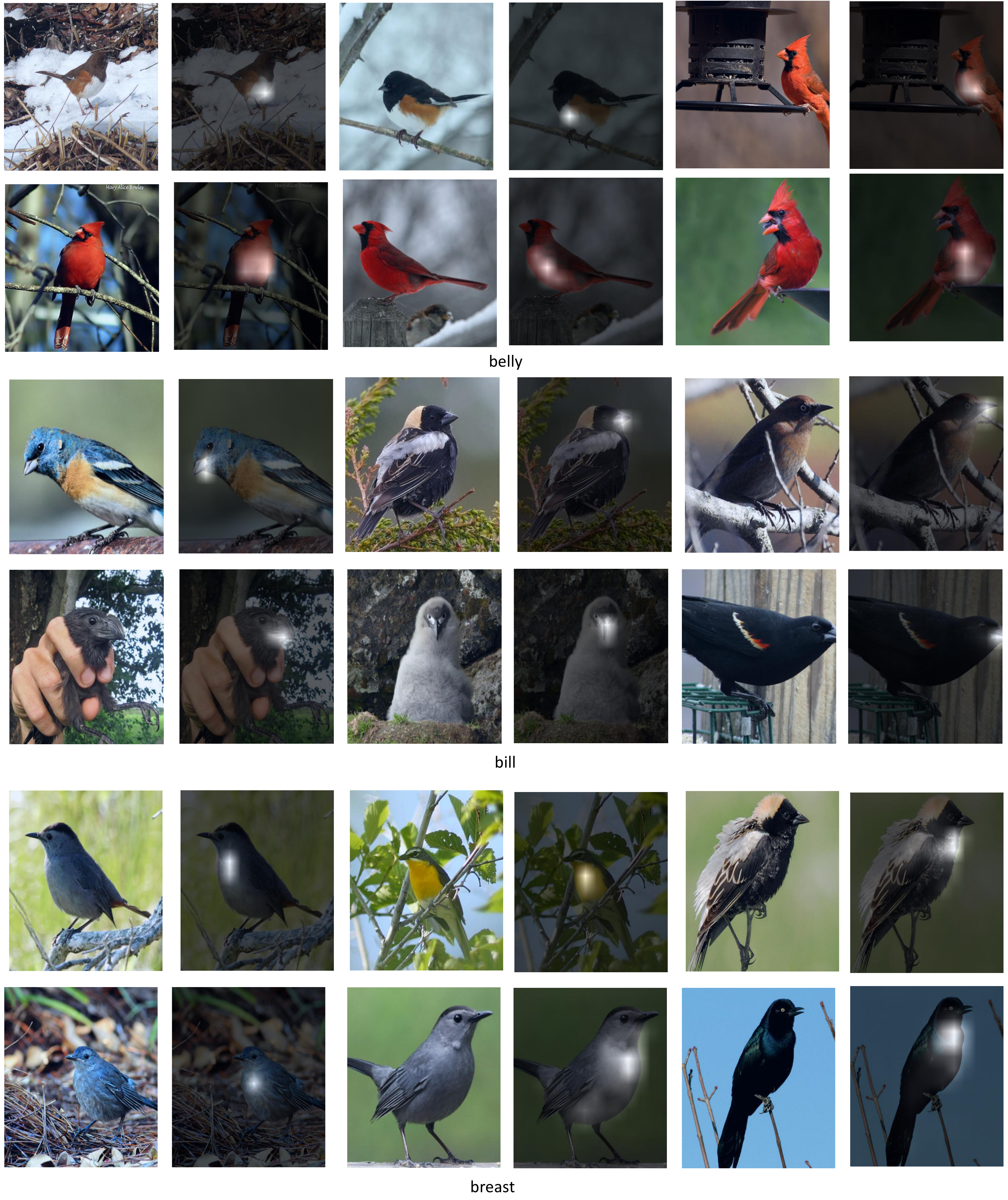}
	\caption{Exemplar images of the CUB bird dataset.}
	\label{fig.GT_error}
	\vspace{-3mm}
\end{figure}

\begin{figure}
	\centering
	\includegraphics[scale=0.45]{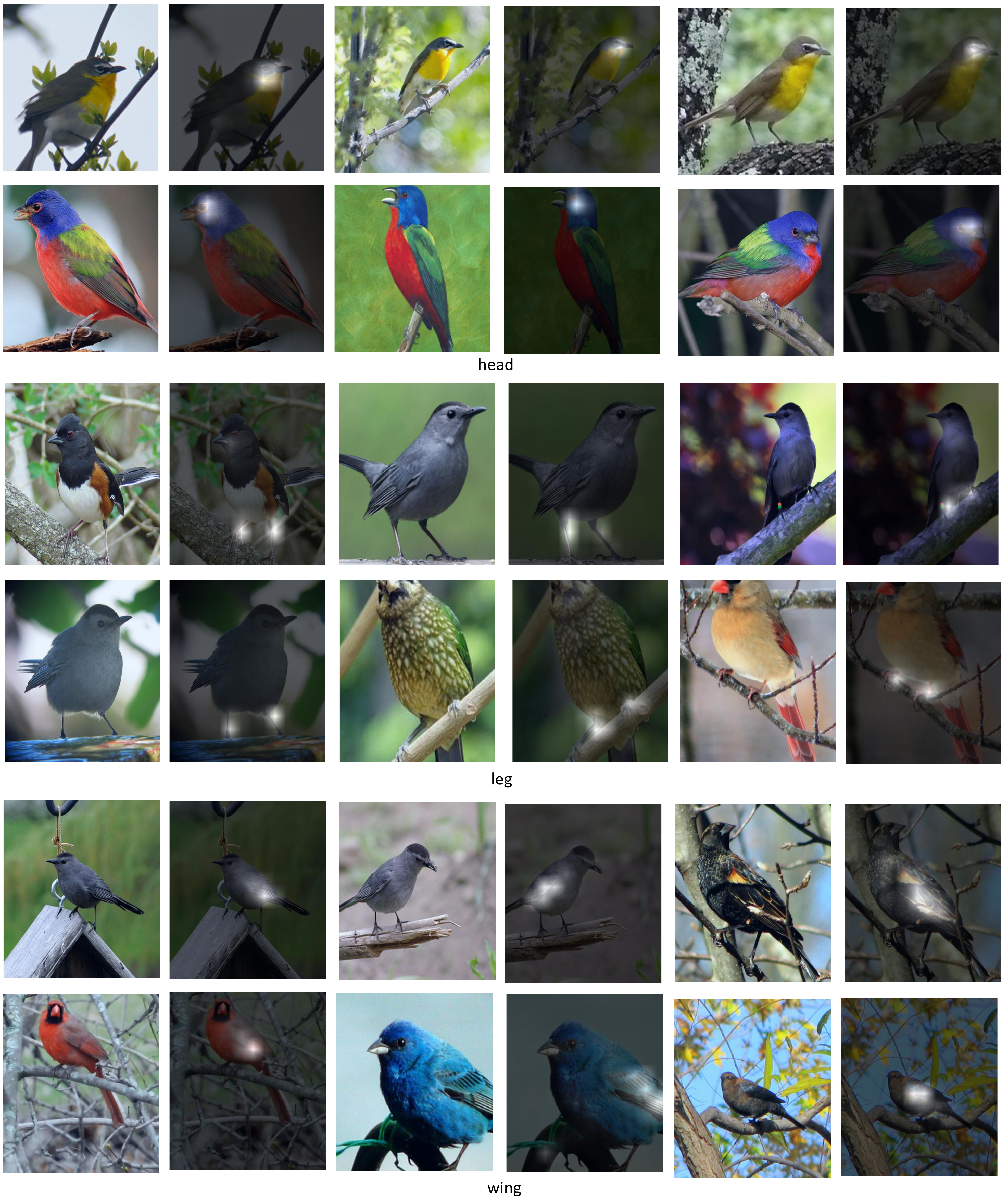}
	\caption{Exemplar images of the CUB bird dataset. }
	\vspace{-3mm}
\end{figure}

\end{document}